\documentclass[conference]{IEEEtran}
\usepackage{amsmath,amsfonts}
\usepackage{algorithm}
\usepackage[noend]{algpseudocode}
\usepackage{algpseudocode}
\usepackage{array}
\usepackage{subfigure}
\usepackage{textcomp}
\usepackage{stfloats}
\usepackage{url}
\usepackage{verbatim}
\usepackage{graphicx}
\usepackage{cite}
\usepackage[dvipsnames]{xcolor}
\usepackage{float}

\usepackage{hyperref}
\hypersetup{
    colorlinks=true,
    linkcolor=blue,
    filecolor=black,      
    urlcolor=blue,
    citecolor=blue,
}

\usepackage[left=1.62cm,right=1.62cm,top=1.9cm]{geometry}
\newcommand{\cmmnt}[1]{}
\begin{document}

\title{Meta Reinforcement Learning for  Strategic IoT Deployments Coverage in Disaster-Response UAV Swarms}   
\author{
    \IEEEauthorblockN{Marwan Dhuheir\IEEEauthorrefmark{1},
    Aiman Erbad\IEEEauthorrefmark{1}, Ala Al-Fuqaha\IEEEauthorrefmark{1}
    }
    \IEEEauthorblockA{\IEEEauthorrefmark{1}Division of Information and Computing Technology, College of Science and Engineering,\\ Hamad Bin Khalifa University, Qatar Foundation, Doha, Qatar.
   }
}
\maketitle

\begin{abstract}
In the past decade, Unmanned Aerial Vehicles (UAVs) have grabbed the attention of researchers in academia and industry for their potential use in critical emergency applications, such as providing wireless services to ground users and collecting data from areas affected by disasters, due to their advantages in terms of maneuverability and movement flexibility. The UAVs' limited resources, energy budget, and strict mission completion time have posed challenges in adopting UAVs for these applications. Our system model considers a UAV swarm that navigates an area collecting data from ground IoT devices focusing on providing better service for strategic locations and allowing UAVs to join and leave the swarm (e.g., for recharging) in a dynamic way. In this work, we introduce an optimization model with the aim of minimizing the total energy consumption and provide the optimal path planning of UAVs under the constraints of minimum completion time and transmit power. The formulated optimization is NP-hard making it not applicable for real-time decision making. Therefore, we introduce a light-weight meta-reinforcement learning solution that can also cope with sudden changes in the environment through fast convergence. We conduct extensive simulations and compare our approach to three state-of-the-art learning models. Our simulation results prove that our introduced approach is better than the three state-of-the-art algorithms in providing coverage to strategic locations with fast convergence. 
\end{abstract}


\section{introduction}
Unmanned aerial vehicles (UAVs) are playing a key role in many applications. Recently, UAVs have been widely used as flying wireless communication base stations due to their advantages over traditional platforms. The UAVs advantages include effective cost, flexibility, easy maneuverability, scalability, and line of sight (LoS) service. These advantages led to the wide adoption of UAVs in critical applications, including surveillance, forest fire detection, object tracking, and wireless communication services in catastrophic disasters (e.g., earthquakes, hurricanes, and floods affected areas) \cite{9456851, padro2019comparison}. Furthermore, UAVs can provide a cost-effective and reliable solution to collect data from dispatched IoT users over a wide geographical area with terrestrial infrastructure impacted by natural or man-made disasters.

UAVs have many advantages over traditional platforms allowing them to effectively accomplish their mission with less energy consumption and within the time constraints \cite{mozaffari2017mobile}. In this way, UAVs can work as data aggregators, which are flying base stations that provide wireless communication services to ground users that cannot be reached with terrestrial base stations. Nevertheless, several challenges need to be addressed to effectively use UAVs, including optimizing mission trajectory, energy consumption, resource allocation, and the completion time of missions.

UAV-based solutions for data collection from IoT devices can optimize the mission's path, energy consumption, and completion time \cite{9031752, 8613833,8908690}. The authors in \cite{9031752} propose to jointly optimize the UAV's positions and the transmit power for reliable information transmission and rapid data collection from ground users. In \cite{8613833}, the authors introduce a joint position and travel path optimization to consume less energy in gathering data. The trajectory path planning optimization aims to collect more user data under energy consumption constraints. 
In \cite{8908690}, the authors used deep reinforcement learning of UAVs' trajectory planning optimization. 
The authors of this study focus on finding the optimal paths of UAVs under time constraints in which the UAVs need to deliver their collected data to a central station before the data becomes irrelevant. All the above-mentioned studies \cite{9031752, 8613833,8908690} focus on optimizing the UAVs path planning to reduce energy and maximizing the coverage to include a larger number of users and collect more data. However, none of these studies investigated planning the UAV paths to cover the affected areas with a focus on the most vulnerable spots to ensure the UAVs plan their trajectories and collect data from the most affected ground IoT devices and upload the data into a remote base stations. Particularly, the UAVs can navigate into the whole area and focus on their trajectories in some strategic locations while maintaining the minimum data rate that guarantees reliable data transmission to other UAVs within the swarm.

Machine learning-based solutions to optimize the path planning of UAVs in dynamic solutions have been addressed in some research studies \cite{10002339, dhuheir2023llhr, 9220821, 8727504}. The works in \cite{10002339, dhuheir2023llhr} investigate using UAVs path optimization to reduce the latency and improve the reliability of the transmission by considering deep reinforcement learning approach. The authors in \cite{9220821} proposed a machine learning solution to provide on-demand services to ground users. In \cite{ 8727504}, the authors use multi-agent reinforcement learning using a Q-learning approach to choose the trajectories of UAVs based on predicting the user movements. Nevertheless, these two studies are not suitable in case of rapid changes in the environment, especially since the UAV path parameters are constantly changing during the mission in dynamic environments. For example, UAVs can join and leave the swarm dynamically for recharging and due to other urgent issues (e.g., damage, malfunction). Due to the complexity of the problem, conventional RL algorithms are slow to converge again to respect the constraints of path planning; hence, we propose meta-learning to quickly adapt neural networks performing in new environments with high efficiency across related tasks \cite{finn2017model}.
 
In this work, we aim to address the problem of UAVs' path planning that provides the minimum energy consumption of UAVs while guaranteeing the minimum data rate is ensured for reliably sending the data to the UAVs by using meta-RL algorithms for coping with dynamic environments through rapid convergence. In our approach, we define strategic locations in the covered geographical area that UAVs need to visit more. These strategic locations represent more affected buildings/neighbourhoods after a disaster or areas with more victims so the UAVs need to plan their trajectory to traverse through these buildings/neighbourhoods more often to ensure timely data collection from severely affected victims. However, our approach is general enough to include wide UAV applications like object tracking and wireless communication services in condensed areas. Our contributions can be summarized as follows:
\begin{itemize}
    \item We develop a system model containing a UAV swarm navigating an area to collect data from IoT devices distributed in the ground. The covered area contains some strategic locations that need better service from the UAV swarm.
    \item We delineate the approach as an optimization problem that seeks to minimize the total energy consumption of the UAVs in the grid while having some constraints ensuring the reliable data delivery among UAVs, minimum energy consumption, optimal path planning with strategic locations.
    \item We investigate the proposed system model through extensive simulations to prove its performance by testing it on various parameters such as changing the number of UAVs in the grid and comparing our meta-RL solution with three state-state-of-art algorithms, namely reinforcement learning with proximal policy optimization (PPO), actor-critic algorithm, and DQN algorithm, and show that our introduced Meta-RL algorithm provides better demand service satisfaction to the strategic locations than the other competitive algorithms.
\end{itemize}
The rest of this article is organized as follows: section \ref{info_system_model} presents the description of our system model. In Section \ref{problem_formulation}, we delineate the problem formulation. Section \ref{Performance_evaluation} explains the implementation results of the proposed approach. At the end, section \ref{conclusion} concludes and discusses future research directions.

\section{system model}
\label{info_system_model}

\begin{figure}[!ht]
    \centering
    \includegraphics[width=0.4\textwidth]{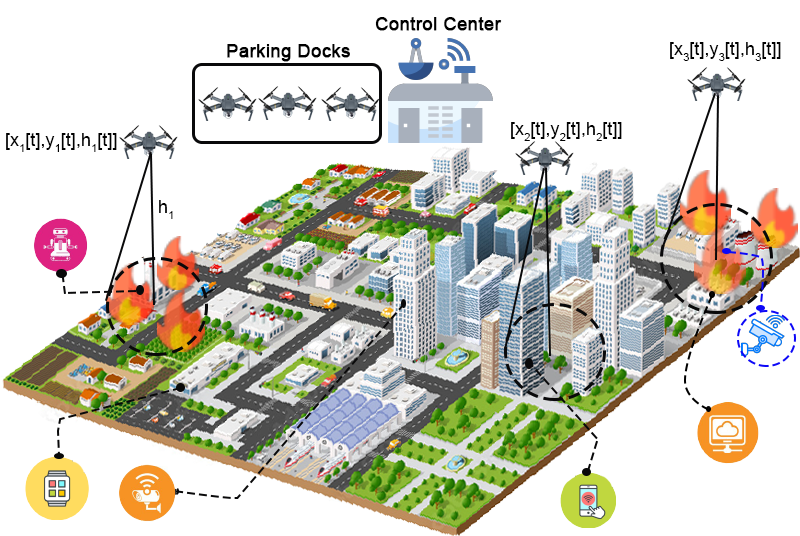}
    \caption{System Model for multi-UAVs covering an area with strategic locations. The UAVs mission is data collection from ground devices.}
    \label{System_Model}
\end{figure}

We divide the covered area into equal-sized cells, in which each cell is monitored by one UAV for data collection from the IoT devices that are distributed in the area as shown in Figure \ref{System_Model}. According to this approach, $U$ UAVs denoted as $u = \{1, 2, \ldots, U\} $ navigate to cover the area focusing on strategic locations in the area. In particular, $U$ UAVs navigate for data collection from $N \gg 1$ ground IoT devices where $i = \{ 1, 2, \ldots, N \}$. The positions of the ground IoT devices are denoted by $i$, $Q_i = [x_i, y_i, 0 ]$, where $i \in N$ are supposed to be known using global positioning systems (GPS). As shown in Figure \ref{System_Model}, the movements of UAVs and their directions are optimized using a ground station (control center). 

Let us suppose that UAVs operate at time frame $T$ where $T > 0$ in seconds (s). The time frame $T$ is divided into small time intervals $t$, such that $0 \leq t \leq T$, thus the 3D positions of UAV $u$ at the specific time instance $t$ can be described as  $Q_u(t) = [x_u(t),y_u(t), h_u(t)] \in \mathbb{R}^3, u \in U$ is a set of UAVs. Moreover, we further assume that the duration time frame $T$ is divided into $M$ equal time slots, determined by the set of $m = \{1, \ldots, M \}$. Each time slot $m$ is defined by the time length of $\mu = \frac{T}{M}$, which is small enough to provide a stable time duration of the UAV 3D location \cite{8255824}. Accordingly, the UAVs’ 3D position can be rewritten as $Q_u[m] = [x_u(t),y_u(t), h_u(t)], m \in M$. The gathered data from ground devices is sent to the ground  base station after the UAVs complete a one-time frame $T$.

\subsection{Wireless Channel Model}

In our approach and for practical scenarios, the obstacles information, including their number, height, and locations, might not be known; hence the randomness of the availability of line of sight (LoS) and non-line of sight (NLoS) channels of the air-to-ground link between UAVs and IoT devices are considered. Note that the LoS and NLoS depend on the type of environment (e.g., rural, urban, suburban, etc.), the location of UAVs and IoT devices, and the altitude of the flying UAVs. Hence, the probability of LoS expression is given by \cite{mozaffari2017mobile}:
\begin{equation}
    P_{ui}^{LoS} = \frac{1}{1+\omega_1 \exp{-\omega_2 [\theta_{ui}-\omega_1]}}
    \label{Los_equation}
\end{equation}
where $\omega_1$ and $\omega_2$ are constant parameters and their values are specified based on the type of the environment, $\theta_{ui}$ represents the elevation angle between the UAV $u$ and the IoT device $i$. Particularly, $\theta = \frac{180}{\pi} \times \sin^{-1}(\frac{h_i}{d_ui})$, where $d_{ui} = \sqrt{(x_i- x_u)^2 +(y_i - y_u)^2 + h_i^2}$ is the distance between the UAV $u$ and the device $i$. The probability of NLoS is given by, $P_{ui}^{NLoS} = 1 - P_{ui}^{LoS}$.

The path loss models of LoS and NLoS are expressed as follows \cite{mozaffari2017mobile}:
\begin{equation}
    L_{LoS}^{ui} = \psi_{LoS}(\frac{4\pi f_c d_{ui}}{c})^2
\end{equation}
\begin{equation}
    L_{NLoS}^{ui} = \psi_{NLoS}(\frac{4\pi f_c d_{ui}}{c})^2
\end{equation}
where $f_c$ is denoted as the carrier frequency and $c$ is the speed of light. $\psi_{LoS}$ and $\psi_{NLoS}$ are denoted to the excessive path loss related to free-space propagation loss for LoS and NLoS, respectively. 

The total average path of the communication link between the UAVs and the IoT devices is given by:
\begin{equation}
    \Bar{L}_{ui} = P_{ui}^{LoS} L_{LoS}^{ui} + P_{ui}^{NLoS} L_{NLoS}^{ui}
\end{equation}\

Moreover, the average channel gain of the communication link between IoT devices and UAVs is $\Bar{g}_{ui} = (1/\Bar{L}_{ui})$. The threshold of minimum data rate for successful transmission is expressed as:
    \begin{equation}
         \rho_{u,i}(lb) = B_{u,i}\: log_{2}(1+\frac{P_r}{\sigma^2})
        \label{data_rate}
    \end{equation}
where $B_{u,i}$ is the transmission bandwidth between the UAV $u$ and IoT device $i$, $P_{r}$ is the received power at the UAV $u$, and $P_{r} =  P_{i} \times \Bar{g}_{ui}$, $P_{i}$ is denoted to the transmit power at device $i$, and $\sigma^2$ represents the thermal noise power. Hence, to achieve a reliable collection time from UAVs from IoT devices, the following condition need to be satisfied:
\begin{equation}
    \rho_{u,i}[m] \geq \rho_{u,i}(lb)
    \label{min_data_rate}
\end{equation}

To satisfy a reliable transmission of data from IoT devices to the flying UAVs, the maximum altitude of UAVs needs to satisfy the minimum SNR $\Gamma$ as follows  \cite{9321452}: 
\begin{equation}
    h_i \leq \bigg ( \frac{P_{max}}{\Gamma F_0^2 \sigma^2 \psi_{LoS}} \bigg )
    \label{max_alitiude}
\end{equation}
where $F_0 =\frac{4\pi f_c}{c}$, and $P_{max}$ is maximum transmit power of each UAV. 
\subsection{Device-to-Device (D2D) Time Delay Model}
D2D time is the time delay to complete the data collection and send it to the central station. The completion time contains the time required to transfer the data from IoT devices $(D_{data})$ and the time required by the UAVs to accomplish the mission $(D_{com})$. The time is given by:
\begin{equation}
    D_{data} = \sum_{i \in N} \frac{K_i}{\rho_{i,k}[M]}
    \label{data_delay}
\end{equation}
\begin{equation}
    D_{com} = \sum_{u \in U} \sum_{m = 1} ^{M} \tau_{u,m}
    \label{cov}
\end{equation}

where $K_i$ is the IoT's data packets, and $\tau_{u,m}$ is the $u^{th}$ UAV's travelling time between two successive time-slots $m$ and $m+1$ which is calculated by:
\begin{equation}
    \tau_{u,m} = \frac{\lVert Q_u[m+1] - Q_u[m] \rVert}{V}, \; m = 1, \ldots , M
\end{equation}

where $V$ is the average speed of $u^{th}$ UAV travelling between two consecutive locations.

Then, the total completion time of $u^{th}$ UAV mission is given by:
\begin{equation}
    D_{tot} = D_{data} + D_{com}
\end{equation}

To satisfy the limitation of completing time, the total completion time should satisfy the following inequalities:
\begin{equation}
    D_{tot} \leq T_{max}
    \label{max_time}
\end{equation}
where $T_{max}$ is the maximum time for UAV to complete its mission.

\subsection{Energy Consumption Model}
The energy consumption of UAV is the total energy consumed by each UAV. The energy consumed by $u^{th}$ UAV can be expressed as:
\begin{equation}
    \varepsilon_u = P_{u,oper} D_{tot} + P_{m,comm} D_{data}
\end{equation}
where $P_{u,oper}$ and $P_{u,comm}$ are the UAV's operating power and UAV's communication power. This energy is designed to calculate the energy consumed by the UAV to follow its planned trajectory and collect data while focusing on strategic locations such that at each time $T$, all strategic locations need to be visited at least one time. Then, the total energy consumed by all the UAVs (in Joules) in the swarm is expressed as:
\begin{equation}
    \varepsilon_{tot} = \sum_{u \in U} \delta_{u,i,q}\; . \; \varepsilon_u
    \label{tot_energy_cons}
\end{equation}
where $\delta_{u,i,q}$ is a binary constraint to encourage UAVs pass through strategic locations and exploit their energy collecting data from the IoT devices, and it is defined as:
\begin{equation}
\small
    \delta_{u,i,q} 
        = \left\{ 
        \begin{array}{l}
            1 \;\text{\;if UAV $u$ is collecting data from IoT $i$ that is}\\
            \text{\; \; in strategic location $q_i$,}\\
            0 \; \text{\;otherwise}\\
        \end{array}
        \right. \quad
\end{equation}

\section{problem formulation}
\label{problem_formulation}
The main objective is to minimize the total energy consumption of UAVs in the swarm presented in equation (\ref{tot_energy_cons}) by finding the optimal positions of UAVs while respecting the maximum completion time, and minimum achievable data rate for reliable data collection. The problem formulation is expressed as follows:
\begin{equation}
\small
    \min_{Q} \; \; \varepsilon_{tot}
    \label{objective_fun}
\end{equation}
\; \; \; \; \; \; \; \; \; \; \; \; \; \; \; s.t. (\ref{min_data_rate}), (\ref{max_alitiude}), (\ref{max_time})
\begin{equation}
    \small
    \forall u \in U,
    \forall q \in Q, 
    \sum_{i=1}^N \delta_{u,i,q} \geq 1
    \label{delta_const}
\end{equation}
\begin{equation}
    \delta_{u,i,q} \in \{0,1\}
    \label{delta_bin}
\end{equation}

The objective function in equation (\ref{objective_fun}) aims to minimize the total energy consumption of UAVs in the swarm by finding the optimal UAVs paths and optimal transmit power while navigating to cover the area, focusing on visiting strategic locations and collecting data from IoT devices. The mission needs to consider some constraints, including the maximum completion time, the minimum data rate, and the maximum altitude for the UAVs for reliable data transmission from IoT devices into the flying UAVs.

The objective function in equation (\ref{objective_fun}) is NP-hard due the non-linearity in constraints (\ref{min_data_rate}), and (\ref{max_alitiude}). Hence, we use meta-reinforcement learning to solve the objective function of the optimization problem and its constraints to provide an efficient sub-optimal solution, which will be described in the next sections.

\section{Meta-reinforcement learning for efficient energy consumption and path planning}

The model handles a dynamic UAV swarm, i.e., at any moment, UAVs can join and leave the swarm (e.g., for recharging); hence, the approach needs to be flexible to these unforeseen changes and can respect the constraints while flying to collect data from the area. Due to the complexity of the problem, conventional RL algorithms are slow to converge again to respect the constraints of path planning; hence, meta-learning is adapted to quickly adapt in new environments with high efficiency across related tasks \cite{finn2017model}. In particular, one model can be trained to learn new tasks more effectively and converge quickly than working correctly with only one single task.

Meta-RL applies meta-learning to reinforcement learning and aims to make the agents learn the general policy of related tasks. A task consists of a set of states and actions, dynamics, and rewards \cite{finn2017model}. In particular, meta-RL agents do not aim to learn the optimal policy of a specific task; instead, they aim to learn the general policy that can be applied to new environments with the same family of tasks. The benefit is that it enables the agents to quickly reach to the optimal policy of new environments with minimum number of episodes.

In our RL approach, one episode contains a set of time steps, and in each step, the algorithm needs to choose the 3D position of each UAV. The decision to select the best position of a UAV in the swarm is abased on several factors, including minimizing the energy consumption, and ensuring the data is sent successfully from the ground IoT device to the UAV. Our RL approach is represented by a Markov Decision Process (MDP) as $(S, A, P, R, \gamma)$, where $S$ is the environment state, $A$ is the action vector which contains two parts (path planning), $P$ is the probability of the possible transition, $R$ represents the rewards that the agent gets for each action, and $\gamma$ is denoted to the discount learning factor.

\begin{enumerate}
    \item \textbf{Environment Modeling}: The environment is a swarm of UAVs navigating the covered area with focusing on strategic locations. Let's define the $\pi$ as the optimal stochastic policy that the agent tries to learn where $\pi: S \times A \rightarrow [0, 1]$. The RL algorithm receives the details from a centralized agent that interacts with the environment, takes action $A$, gets either reward/penalty of that action, and reaches the optimal policy $\pi^*$ by the received reward $R$ at each time step $t$. The algorithm tries to reach to the optimal policy $\pi^*$ with the maximum $v^{\pi^*} (s)$ for all parameters $s \in S$. The value set of $v^{\pi}(s)$ is denoted to the feedback from the reward after implementing action $a$ on state $s$, where $\mathbf{E}$ is denoted to the expected value and it can be illustrated by:
    \begin{equation}
    \small
        v^{\pi}(s) = \mathbf{E}_{a_{t},s_{t+1}}\Bigg(\sum_{k=1}^{\infty} \gamma^{k-t} R_k|S_t = s\Bigg),
        \label{12th_eqn}
    \end{equation}
    \item \textbf{States and Actions}: The RL agents needs helpful hints about the environment to improve the system performance. The state vector in our approach consists of the positions of UAVs and strategic locations and the visited cells in the grid. The action of the agent is the UAV direction focusing on strategic locations.
    \item \textbf{Reward function}: The reward is a crucial factor to help the algorithm learn the optimal policy. In our algorithm, the agent is positively rewarded when UAV’s direction is chosen such that it visited one of the strategic locations and consumes less energy. The agent gets a negative reward if the agent's next UAV position is higher than the minimum derived data rate or collides with other UAVs. The UAVs need to pass through strategic locations and collect data from the IoT devices. Particularly, each strategic location has different demand services based on its importance that UAVs need to pass through and satisfy part of this demand service factor which represents the quality of services (QoS) in these strategic locations. When UAVs pass through these strategic locations, it satisfies part of the demand service denoted by $\phi_\Omega$. Therefore, we define the rewards of satisfying the number of visits to these strategic locations as done in \cite{10002339} by:
    \begin{equation}
    \small
    R_{\Omega} = \frac{1}{1+\sum_{i=1}^{\Omega} \phi_\Omega (t)}
\label{Qos_eqn}
\end{equation}

\end{enumerate}

Algorithm \ref{actor_critic} delineates the steps of meta-RL algorithm for converging and respecting the constraints related to the objective function in (\ref{objective_fun}). According to this algorithm, if the UAV does not collide with other UAVs, it gets a reward (lines 10-12). Then, if the UAV chooses the next step producing less energy, above the limit of minimum data rate and less than the maximum expected time completion, it also gets a reward. The agent will get an extra reward if it visits one strategic location at that time, encouraging UAVS to collect data from the strategic locations (lines 13-16).
\begin{algorithm}
\caption{Meta-RL}
\label{actor_critic}
\begin{algorithmic}[1]
\small
\State \text{Q-network parameters initialization $\theta$ and $\theta_v$.}
\State \text{old Q-network parameters initialization $\theta^{\prime}$ and $\theta_v^{\prime}$}
\For{each episode $r$ $\in$ $R$}
\State \text{gradient initialization: $d\theta \leftarrow 0$ and $d\theta_v \leftarrow 0$}
\State \text{Q-network initialization: $\theta^{\prime} = \theta$ and $\theta_v^{\prime} =\theta_v$}
\For{each $t$ $\in$ $T$}
\For{each $u \in \{1,2,3,...,U\}$}
    \State \text{$S(u) = \{S_u[m], S_{SL}[m]\}$}
    \State \text{choose action $A_i$ based on $\epsilon$}
    \If{$ Q_u[t] \notin Q_u [T]$ and (\ref{min_data_rate}), (\ref{max_alitiude}), (\ref{max_time})} \qquad
    \State \text{$R_t = R_t + 1 \qquad$}
    \EndIf
    \If{$S_u \in StrategicLosaction$}
    \State \text{$energyEpisode \; += \; \varepsilon_{tot} (Q, N_i)$ equation (\ref{objective_fun})} 
    \State \text{$R_t = R_t + R_{\Omega} \qquad$ equation (\ref{Qos_eqn})} 
    \EndIf
    \State \text{observe $S_{i+1}$}
    \State \text{observe $R_t$}
    \State \text{system reset with new UAVs positions}
    \State \text{save $(S_i, A_i,r_i,S_{i+1})$ in replay memory}
    \State \text{taking a minibatch of $(S_i, A_i,r_i,S_{i+1})$}
    \State \text{gradient accumulation wrt$\theta^{\prime}: d\theta \leftarrow d\theta +$}
    \text{\qquad \qquad \qquad \qquad \quad \quad $\nabla_{\theta^{\prime}}\log\pi(a_i|s_i;\theta^{\prime})(R-V(s_i;\theta_v^{\prime}))$}
    \State \text{gradient accumulation wrt$\theta_v^{\prime}: d\theta \leftarrow d\theta +$}
    \text{\qquad \qquad \qquad \qquad \quad \quad $\partial(R-V(s_i;\theta_v^{\prime}))^2 / \partial\theta_v^{\prime}$}  
\EndFor
\EndFor
\State \text{asynchronous update of $\theta$ using $d\theta$ and $\theta_v$ using $d\theta_v$}
\EndFor
\end{algorithmic}
\end{algorithm}
\section{simulation results and analysis}
\label{Performance_evaluation}

In our approach, we used 440 m $\times$ 440 m with 25 equal-sized cells. Among these 25 cells, we have three strategic locations located in different places on the grid. For environment type, we used urban area with parameters of $\omega_1$ and $\omega_2$ as 11.95 and 0.14, respectively. The $\psi_{LoS}$ and $\psi_{NLoS}$ are 3 dB and 23 dB, respectively. The list of parameters is summarized in table \ref{calculation_data_summary}.
\begin{table}[!ht]
    \caption{Simulation Parameters.}
    \label{calculation_data_summary}
    \centering
    \begin{tabular}{|p{3cm}|p{3cm}|p{1.5cm}|}\hline
        Parameters & Description & Value \\\hline
        max transmit power & $P_{max}$ & 200 mW \\ \hline
        $\sigma^2$ & noise power & -170 dBm \\ \hline
       $\gamma$ & Discount factor &  0.85\\\hline
       $\alpha$ & Learning rate & 0.0001 \\ \hline
       $B_u$ & Bandwidth &  1 MHz \\ \hline
       $K_i$ & IoT’s data packet size & 1 Mbits \\ \hline
       $V$ & average ﬂight speed & 10 m/s  \\ \hline
        operating power & $P^{oper}_u$ & 300W \\ \hline
        communication power & $P^{comm}_u$ & 5W\\ \hline
    \end{tabular}
\end{table}

\begin{figure}[!ht]
    \centering  \includegraphics[width=0.4\textwidth]{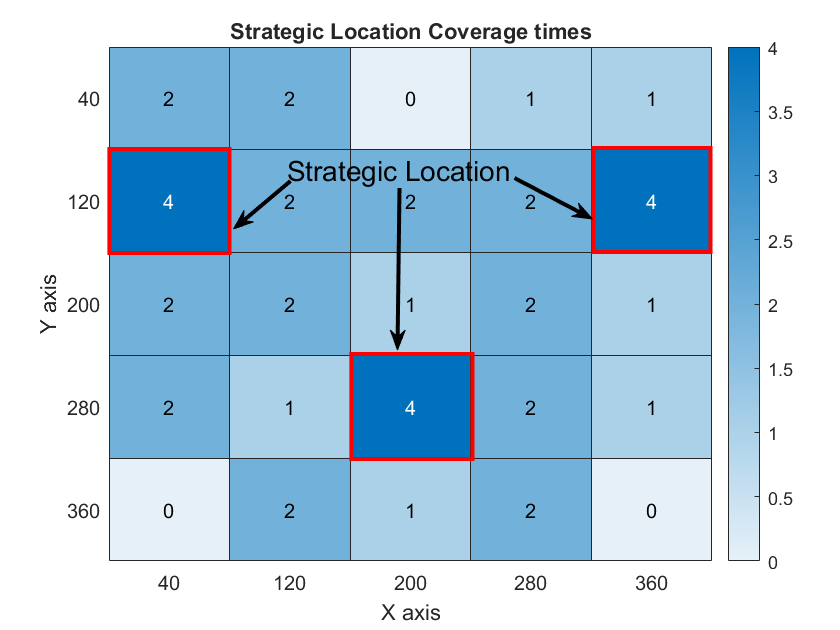}
    \caption{The number of visitations to strategic locations in one time frame $T$.}
    \label{fig:heatmab_fig}
\end{figure}
Figure \ref{fig:heatmab_fig} expresses the number of visitations to strategic locations compared to non-strategic locations in a one-time frame $T$. As shown in the figure, our approach can successfully encourage the UAVS to pass through these strategic locations more often than the other locations.
\begin{figure}[!ht]
    \centering    
    \includegraphics[width=0.4\textwidth]{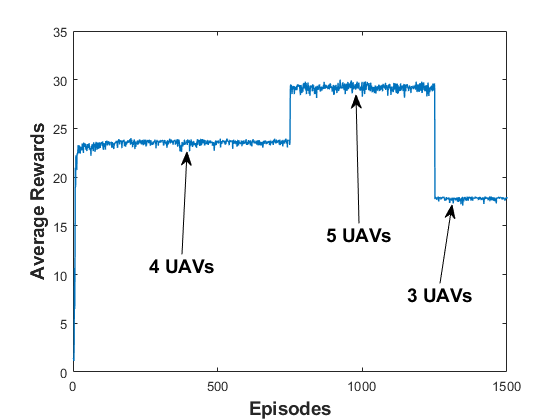}
    \caption{Adaptivity of Meta-RL algorithm to the environment changes of the learning. The algorithm started with 4 UAVs, then 1 more UAV joined the swarm, and after that, 2 UAVs left the swarm. Meta-RL algorithm learns the optimal policy quickly and converges to its maximum expected reward.}
    \label{fig:Env_changes}
\end{figure}

We investigate a practical scenario when the number of UAVs joins and disconnects the swarm during the learning. Figure \ref{fig:Env_changes} shows the related implementation of the average reward convergence. The simulation started with a swarm of four UAVs, and we highlight that Meta-RL converges rapidly to the maximum expected reward, and it takes 900 episodes to converge into the maximum expected reward. During the learning, one UAV joined the swarm, and Meta-RL can adapt to the new changes in the environment and converge very quickly to the maximum expected reward. After that, two UAVs depart the swarm (e.g., for recharging), and Meta-RL can converge to the maximum expected reward.

\begin{figure*}[!t]
\centering
	\mbox{
	    \hspace{-7mm} \subfigure[\label{fig:STR_Energy}]{\includegraphics[scale=0.20]{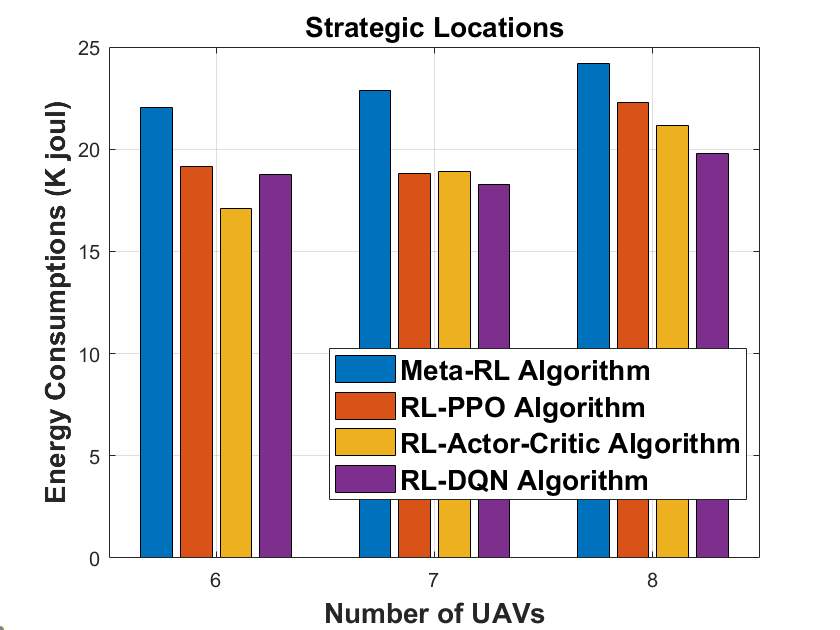}
	   }
	     \hspace{-7mm}
	     \subfigure[\label{fig:non_STR_Energy}]{\includegraphics[scale=0.20]{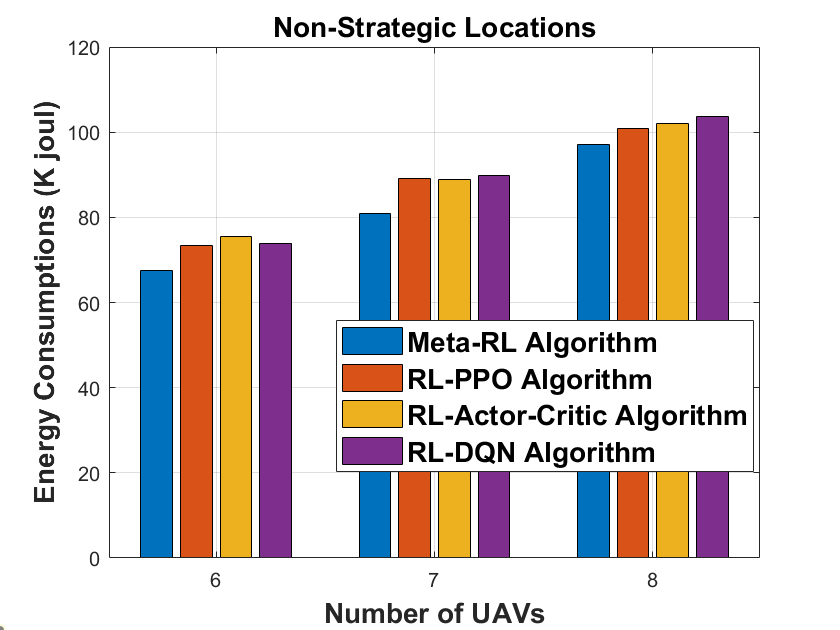}}
             \hspace{-7mm}
	     \subfigure[\label{fig:convergence_speed}]{\includegraphics[scale=0.20]{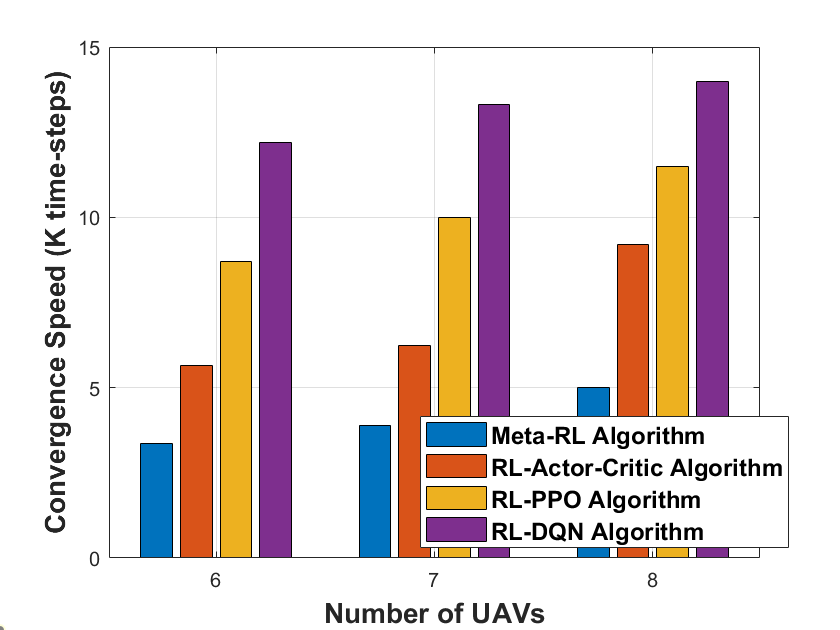}}
      \hspace{-7mm}
	     \subfigure[\label{fig:QoS_algorithms_comparison}]{\includegraphics[scale=0.20]{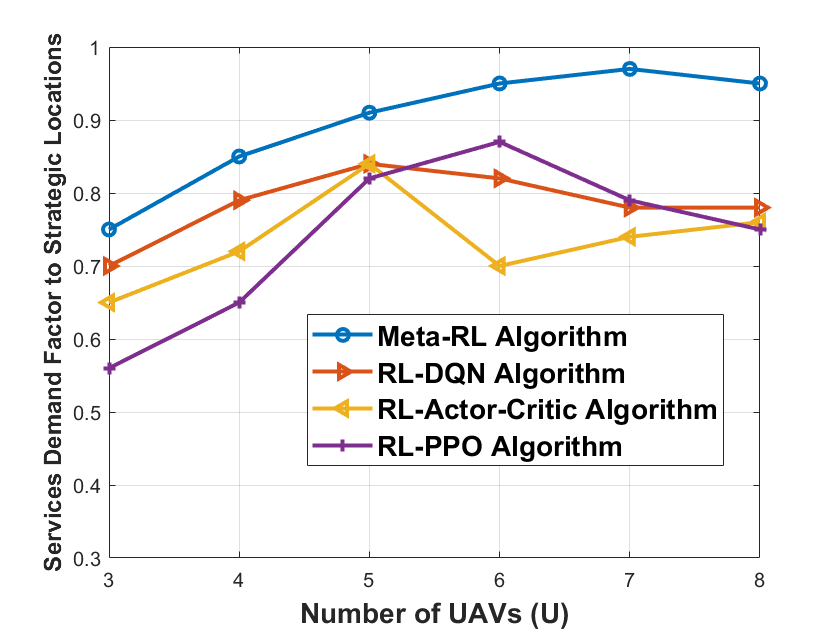}}
      } 
	\caption{Energy consumption of different algorithms in terms of strategic locations and no-strategic locations, convergence speed, and demand service satisfaction.}
	\label{fig:Total_comparsion}
\end{figure*}

Figures \ref{fig:STR_Energy} and \ref{fig:non_STR_Energy} compare the Meta-RL solution and traditional RL algorithms in case of energy consumption in strategic and non-strategic locations when the number of UAVs changes in the swarm. As shown in the figures, Meta-RL tends to spend more energy than others on strategic locations and lower energy consumption on non-strategic locations as its priority is to serve strategic locations. Nevertheless, Meta-RL outperforms the traditional RL algorithms in terms of successfully satisfying the demand service needs of strategic locations, reaching 96\% when 7 UAVs are used, as shown in Figure \ref{fig:convergence_speed}. Meta-RL also outperforms the traditional RL algorithms regarding convergence speed, as shown in Figure \ref{fig:QoS_algorithms_comparison}. The Meta-RL algorithm needs fewer episodes than the other algorithms to learn the optimal policy and hence the ability to learn faster.   

\section{conclusion}
\label{conclusion}
In this article, we investigated the optimization of path planning and energy consumption of UAV swarms in the task of collecting data from IoT devices while focusing on strategic locations to visit more often than the other areas in the grid. We formulated the problem as an optimization problem that seeks the minimum energy consumption by controlling the UAV deployments and ensuring the minimum data rate. We propose an efficient sub-optimal solution using meta-reinforcement learning for dealing with dynamic environments and for fast convergence. We compared our proposed solution with three competitive solutions and showed the effective of our approach in providing better service demand in strategic locations. Our simulation results demonstrated the effectiveness of the proposed method in terms of environment adaptive to sudden changes and fast convergence into the maximum rewards. In future work, we plan to study the interference and the Doppler effect on UAV movements.
\section*{Acknowledgment}
This work was made possible by NPRP grant \# NPRP13S-0205-200265 from the Qatar National Research Fund (a member of Qatar Foundation). The findings achieved herein are solely the responsibility of the authors.
\bibliographystyle{IEEEtran}
\bibliography{bibliography.bib}

\begin{thebibliography}{10}
\providecommand{\url}[1]{#1}
\csname url@samestyle\endcsname
\providecommand{\newblock}{\relax}
\providecommand{\bibinfo}[2]{#2}
\providecommand{\BIBentrySTDinterwordspacing}{\spaceskip=0pt\relax}
\providecommand{\BIBentryALTinterwordstretchfactor}{4}
\providecommand{\BIBentryALTinterwordspacing}{\spaceskip=\fontdimen2\font plus
\BIBentryALTinterwordstretchfactor\fontdimen3\font minus
  \fontdimen4\font\relax}
\providecommand{\BIBforeignlanguage}[2]{{%
\expandafter\ifx\csname l@#1\endcsname\relax
\typeout{** WARNING: IEEEtran.bst: No hyphenation pattern has been}%
\typeout{** loaded for the language `#1'. Using the pattern for}%
\typeout{** the default language instead.}%
\else
\language=\csname l@#1\endcsname
\fi
#2}}
\providecommand{\BIBdecl}{\relax}
\BIBdecl

\bibitem{9456851}
Q.~Wu, J.~Xu, Y.~Zeng, D.~W.~K. Ng, N.~Al-Dhahir, R.~Schober, and A.~L.
  Swindlehurst, ``A comprehensive overview on 5g-and-beyond networks with uavs:
  From communications to sensing and intelligence,'' \emph{IEEE Journal on
  Selected Areas in Communications}, vol.~39, no.~10, pp. 2912--2945, 2021.

\bibitem{padro2019comparison}
J.-C. Padr{\'o}, F.-J. Mu{\~n}oz, J.~Planas, and X.~Pons, ``Comparison of four
  uav georeferencing methods for environmental monitoring purposes focusing on
  the combined use with airborne and satellite remote sensing platforms,''
  \emph{International journal of applied earth observation and geoinformation},
  vol.~75, pp. 130--140, 2019.

\bibitem{mozaffari2017mobile}
M.~Mozaffari, W.~Saad, M.~Bennis, and M.~Debbah, ``Mobile unmanned aerial
  vehicles (uavs) for energy-efficient internet of things communications,''
  \emph{IEEE Transactions on Wireless Communications}, vol.~16, no.~11, pp.
  7574--7589, 2017.

\bibitem{9031752}
Z.~Huang, C.~Chen, and M.~Pan, ``Multiobjective uav path planning for emergency
  information collection and transmission,'' \emph{IEEE Internet of Things
  Journal}, vol.~7, no.~8, pp. 6993--7009, 2020.

\bibitem{8613833}
M.~B. Ghorbel, D.~Rodríguez-Duarte, H.~Ghazzai, M.~J. Hossain, and H.~Menouar,
  ``Joint position and travel path optimization for energy efficient wireless
  data gathering using unmanned aerial vehicles,'' \emph{IEEE Transactions on
  Vehicular Technology}, vol.~68, no.~3, pp. 2165--2175, 2019.

\bibitem{8908690}
S.~Wan, J.~Lu, P.~Fan, and K.~B. Letaief, ``Toward big data processing in iot:
  Path planning and resource management of uav base stations in mobile-edge
  computing system,'' \emph{IEEE Internet of Things Journal}, vol.~7, no.~7,
  pp. 5995--6009, 2020.

\bibitem{10002339}
M.~A. Dhuheir, E.~Baccour, A.~Erbad, S.~S. Al-Obaidi, and M.~Hamdi, ``Deep
  reinforcement learning for trajectory path planning and distributed inference
  in resource-constrained uav swarms,'' \emph{IEEE Internet of Things Journal},
  vol.~10, no.~9, pp. 8185--8201, 2023.

\bibitem{dhuheir2023llhr}
M.~Dhuheir, A.~Erbad, and S.~Sabeeh, ``Llhr: Low latency and high reliability
  cnn distributed inference for resource-constrained uav swarms,'' in
  \emph{2023 IEEE Wireless Communications and Networking Conference
  (WCNC)}.\hskip 1em plus 0.5em minus 0.4em\relax IEEE, 2023, pp. 1--6.

\bibitem{9220821}
Q.~Zhang, W.~Saad, M.~Bennis, X.~Lu, M.~Debbah, and W.~Zuo, ``Predictive
  deployment of uav base stations in wireless networks: Machine learning meets
  contract theory,'' \emph{IEEE Transactions on Wireless Communications},
  vol.~20, no.~1, pp. 637--652, 2021.

\bibitem{8727504}
X.~Liu, Y.~Liu, Y.~Chen, and L.~Hanzo, ``Trajectory design and power control
  for multi-uav assisted wireless networks: A machine learning approach,''
  \emph{IEEE Transactions on Vehicular Technology}, vol.~68, no.~8, pp.
  7957--7969, 2019.

\bibitem{finn2017model}
C.~Finn, P.~Abbeel, and S.~Levine, ``Model-agnostic meta-learning for fast
  adaptation of deep networks,'' in \emph{International conference on machine
  learning}.\hskip 1em plus 0.5em minus 0.4em\relax PMLR, 2017, pp. 1126--1135.

\bibitem{8255824}
Y.~Zeng, X.~Xu, and R.~Zhang, ``Trajectory design for completion time
  minimization in uav-enabled multicasting,'' \emph{IEEE Transactions on
  Wireless Communications}, vol.~17, no.~4, pp. 2233--2246, 2018.

\bibitem{9321452}
K.~Chen, Y.~Wang, J.~Zhao, X.~Wang, and Z.~Fei, ``Urllc-oriented joint power
  control and resource allocation in uav-assisted networks,'' \emph{IEEE
  Internet of Things Journal}, vol.~8, no.~12, pp. 10\,103--10\,116, 2021.

\end{thebibliography}
\end{document}